\documentclass[letterpaper, 10 pt, conference]{ieeeconf}
\IEEEoverridecommandlockouts
\newcommand{\eg}{\textit{e.g.,}~} %
\newcommand{\ie}{\textit{i.e.,}~} %
\newcommand{\naive}{na\"ive }      %
\let\labelindent\relax %<-this prevents a conflict with IEEE \labelindent when using enumitem
\usepackage[inline]{enumitem}\setlist[enumerate]*{label=(\roman*)}

\newcommand*{\sref}[1]{\S\ref{s:#1}}            % section
\usepackage[pdftex]{graphicx}

\usepackage[font=footnotesize]{subcaption}
\usepackage[font=footnotesize]{caption}

\usepackage[nobreak]{cite}

\usepackage[ruled,linesnumbered]{algorithm2e}
\SetAlCapFnt{\footnotesize}
\SetAlgorithmName{Alg.}{Alg.}{List of Algorithms}

\usepackage{cancel} % cancel slash
\usepackage{times} % times symbol
\usepackage{amssymb, amsmath} % additional math functions
\newcommand{\argmin}{\mathop{\rm arg~min}\limits} % math form arg min

\newcommand\figref[1]{\text{Fig.~\ref{fig:#1}}}
\newcommand\equref[1]{\text{Eq.~(\ref{#1})}}
\newcommand\algorithmref[1]{\text{Alg.~\ref{algorithm:#1}}}

\usepackage{url}

\title{\LARGE \bf
Disturbance--injected Robust Imitation Learning with Task Achievement
}
\author{Hirotaka Tahara$^{1}$, Hikaru Sasaki$^{1}$, Hanbit Oh$^{1}$, Brendan Michael$^{1}$, and Takamitsu Matsubara$^{1}$% <-this % stops a space
\thanks{  
$^{1}$All authors are with the Division of Information Science, Graduate School of Science and Technology, Nara Institute of Science and Technology, Japan.
}%
\thanks{This work is supported by JST [Moonshot Research and Development], Grant Number [JPMJMS2032].}%
}

\begin{document}
\maketitle
\thispagestyle{empty}
\pagestyle{empty}

%%%%%%%%%%%%%%%%%%%%%%%%%%%%%%%%%%%%%%%%%%%%%%%%%%%%%%%%%%%%%%%%%%%%%%%%%%%%%%%%
\begin{abstract}
Robust imitation learning using disturbance injections overcomes issues of limited variation in demonstrations. However, these methods assume demonstrations are optimal, and that policy stabilization can be learned via simple augmentations. In real-world scenarios, demonstrations are often of diverse-quality, and disturbance injection instead learns sub-optimal policies that fail to replicate desired behavior. To address this issue, this paper proposes a novel imitation learning framework that combines both policy robustification and optimal demonstration learning. Specifically, this combinatorial approach forces policy learning and disturbance injection optimization to focus on mainly learning from high task achievement demonstrations, while utilizing low achievement ones to decrease the number of samples needed. The effectiveness of the proposed method is verified through experiments using an excavation task in both simulations and a real robot, resulting in high-achieving policies that are more stable and robust to diverse-quality demonstrations. In addition, this method utilizes all of the weighted sub-optimal demonstrations without eliminating them, resulting in practical data efficiency benefits.
\end{abstract}

%%%%%%%%%%%%%%%%%%%%%%%%%%%%%%%%%%%%%%%%%%%%%%%%%%%%%%%%%%%%%%%%%%%%%%%%%%%%%%%%
\section{INTRODUCTION}

\begin{figure}[tb]
\centering
\includegraphics[width=1.0\hsize] {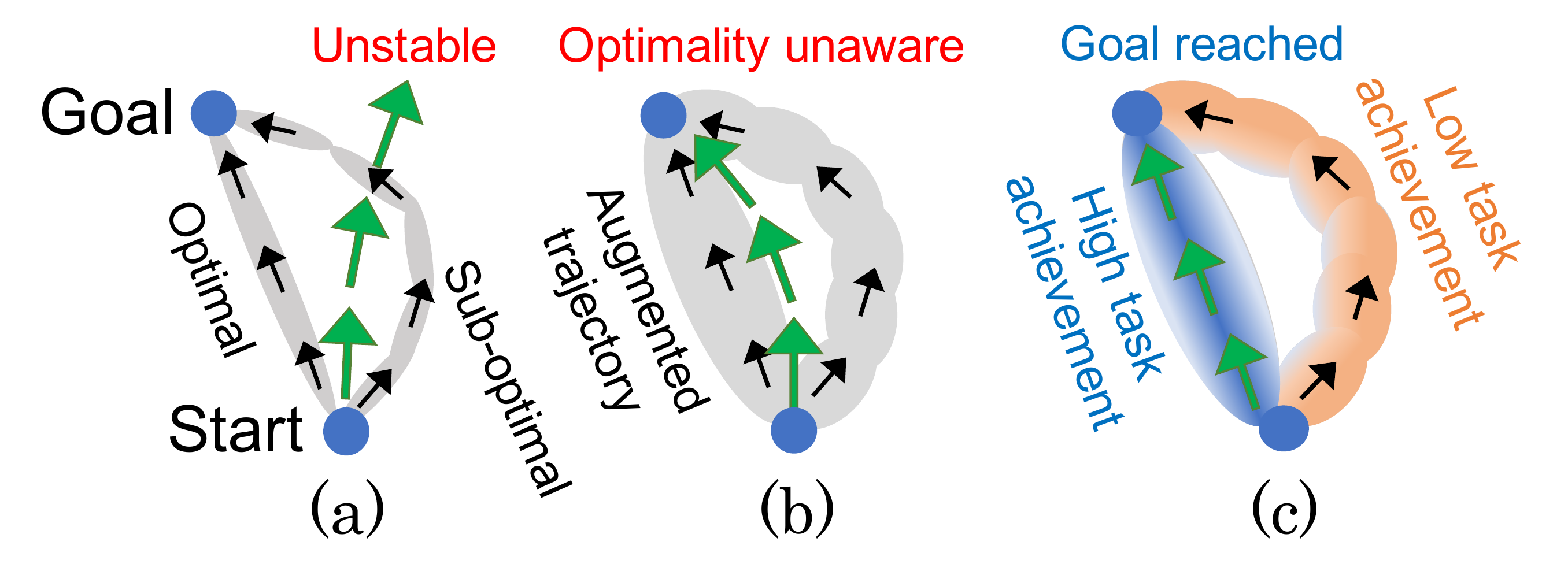}
\caption{ 
Reaching task by imitation learning. (a) A demonstrator gives optimal and sub-optimal demonstrations (black arrows). Learned policy causes compounding errors (green arrows). (b)  Augmentation of action distribution (gray shaded) mitigates the error. However, the optimality of demonstrations is not considered, leading to sub-optimal policy learning. (c) Combining task achievement weighting with robustification enables robust, optimal policy learning.
}
\label{fig:method_proposed_framewark}
\end{figure}

Behavior Cloning (BC) \cite{pomerleau1991efficient} is widely used in robotics as an imitation learning (IL) method \cite{zhang2018deep, osa2018algorithmic} to leverage human demonstrations for learning control policies. Learning from humans is particularly desirable from the perspective of safe interactive robot control, as learned policies are based on the demonstrator's behavior, and requires few samples for training \cite{osa2018algorithmic}.
A common issue affecting learning via behavior cloning is \textit{limited variation} in demonstration data, resulting in overly-specific, poorly generalized policies that are not robust to deviations in behavior. Specifically, learned policies may be influenced by error compounding (also known as covariate shift \cite{ross2010efficient}), where there arises a mismatch between the distributions of data used for training and testing. To robustify learning, stochastic perturbations, known as \textit{disturbance injections}, are added to the demonstrator's actions, augmenting the learning space and resulting in stable policies \cite{laskey2017dart}. However, a key limitation that restricts real-world control, is the assumption that demonstrators are proficient in the task and can provide consistently high-quality demonstrations. Specifically, a key assumption is that error compounding can be solved by assuming demonstration data is homogeneous, and can be used to learn an optimal policy simply by learning flexible generalizations of the demonstration data. In real-world scenarios, data is often heterogeneous and of varying quality, due to the difficulty of the task or human inexperience \cite{grollman2011donut, wu2019imitation, hamaya2020learning}. In addition, demonstrators may perform idiosyncratic behavior, which might not be task optimal (\eg unintentional drifting \cite{coates2008learning}), resulting in \textit{diverse-quality demonstrations}. In this scenario, \naive application of disturbance injections does not consider the demonstration quality, and this diverse-quality bias policy learning, leading to over-generalized policies. An example of this problem is shown in \figref{method_proposed_framewark}.

To limit the impact of poor quality demonstrations, this paper takes influence from reinforcement learning \cite{peters2007reinforcement} to learn policies by weighting the contribution of demonstrations, proposed as \textit{task achievement weighted disturbance injections (TAW-DI)}. Specifically, maximizing the \textit{task achievement} is applied to motivate the policy learning method to selectively update based on high-quality demonstrations, and dissuade from learning from poor ones. Importantly, utilizing weighted sub-optimal trajectories, which may contain both non-optimal and optimal parts, contributes to accelerated convergence of the policy performance. This is particularly appealing, as collecting data for IL is demanding, due to task complexity and limitations on demonstrator availability.

Specifically, an iterative process is proposed: 
\begin{enumerate*}
\item disturbance injected augmented trajectories are generated to minimize task achievement weighted covariate shift and,
\item the policy is updated using the task achievement weighted trajectories, to concentrate trajectories around the space of those with high task achievement.
\end{enumerate*}

Evaluation results show the framework learns robust policies undeterred by either limited variation or diverse-quality of demonstrations, through robotic excavation tasks in both simulations and a scale excavator, and outperforms methods that explicitly account for these issues independently.

%%%%%%%%%%%%%%%%%%%%%%%%%%%%%%%%%%%%%%%%%%%%%%%%%%%%%%%%%%%%%%%%%%%%%%%%%%%%%%%%
\section{RELATED WORK}
The proposed method combines features of robust imitation learning and reward weighting inspired by reinforcement learning. As such, both topics are discussed in the context of motivating the proposed combinatorial approach.

\subsection{Robust Imitation Learning}
Policy learning via imitation of demonstrator behavior such as Behavior cloning (BC) \cite{pomerleau1991efficient} is widely used in robot control \cite{zhang2018deep, osa2018algorithmic}. However, BC suffers from \textit{covariate shift} \cite{ross2010efficient}, where policies trained on a limited variation of demonstrations, fail to control correctly for predicted actions that diverge from the strict expected distribution observed during training. To address this problem, data augmentation methods such as dataset aggregation (DAgger\cite{ross2011reduction}) or disturbance injection (DART\cite{laskey2017dart}) can reduce covariate shift and robustify the policy around the augmented demonstration data.

While robustification mitigates distribution mismatch, there is no guarantee demonstrators perform optimally. Diverse-quality demonstrations are common in real applications with human demonstrations \cite{grollman2011donut, wu2019imitation, hamaya2020learning}. Attempting to apply robust IL to remove this error, results in learning policies concentrate around the average of the entire demonstration data. This quality agnostic approach is undesirable, as while it prevents error compounding, the resulting dataset-centered policy is clearly different than the optimal.

\subsection{Reward weighting}
\label{s:rl}
Diverse-quality demonstrations are antithetical to the assumptions of traditional robust IL methods, which assume high-quality, consistent demonstrations. In contrast, reinforcement learning (RL) does not assume the availability of a demonstrator, and policies are learned via trial-and-error exploration and exploitation of rewards (also known as the task achievement \cite{kinose2020integration}) from the environment, resulting in successful autonomous learning \cite{peters2007reinforcement, arulkumaran2017deep, tangkaratt2020variational}.

On the other hand, RL expects immediate rewards, but is unsuitable to episodic task achievement where we only care about the end result. Such a sparse task achievement cannot evaluate the random actions of RL; therefore, the learning for policy improvement becomes inefficient \cite{goecks2020integrating}. In contrast, our method uses demonstrated actions instead of random actions, and at least the task will be accomplished by demonstrators. This enables wide applicability to other tasks where a specific indicator can evaluate the task achievement at the end of the task (\eg pick and place task). In addition, learning a policy model via trial and error in RL requires a large number of action steps to collect data over the entire environment, and this exploration phase may be unsafe for robot control \cite{garcia2015comprehensive}. This property is limiting the applicability of RL for real-world scenarios.

\subsection{Hybridisation: IL with reward weighting}
While direct application of RL is problematic, reward weighting \cite{peters2007reinforcement} can be utilized to address the issue of diverse-quality demonstrations. Prior work has explored combining IL and RL, for example, using IL as a starting policy in the exploration phase of RL to speed up performance convergence \cite{kober2010imitation, goecks2020integrating}. Additionally, recent studies \cite{brown2020better, chen2020learning} tackle the issue of diverse-quality demonstrations by injecting disturbances into learned IL policies and using estimated rewards. However, utilizing the IL policy, which is in the learning process and is not robustified against covariate shift, can be potentially dangerous for a real robot. Furthermore, trial-and-error exploration in RL accelerates this risk. To address this fundamental problem, robustification of policy learning with DART, which requires only the safe demonstrator's policy, is significant and should be incorporated into IL-based frameworks.

%%%%%%%%%%%%%%%%%%%%%%%%%%%%%%%%%%%%%%%%%%%%%%%%%%%%%%%%%%%%%%%%%%%%%%%%%%%%%%%%
\section{PRELIMINARIES}

\subsection{Behavior Cloning (BC)}
The objective of IL is to learn policies that replicate a demonstrator's behavior, using demonstration trajectories $\boldsymbol{\tau}$ consisting of sequences of states $\mathbf{s}_t$ and actions $\mathbf{a}_t$, where $\boldsymbol{\tau}=\left(\mathbf{s}_{1}, \mathbf{a}_{1}, \mathbf{s}_{2}, \mathbf{a}_{2}, \ldots, \mathbf{a}_{T-1}, \mathbf{s}_T\right)$, and $T$ is a total steps of a trajectory. The trajectory distribution associated with the dynamics $p(\mathbf{s}_{t+1} | \mathbf{s}_t, \mathbf{a}_t)$ and the parametric policy $\pi_{\theta}(\mathbf{a}_t | \mathbf{s}_t)$ with a parameter $\theta$ is defined as:
\begin{align}
p(\boldsymbol{\tau} | \pi_{\theta} ) = p(\mathbf{s}_1) \prod_{t=1}^{T} \pi_{\theta}(\mathbf{a}_t | \mathbf{s}_t) p(\mathbf{s}_{t+1} | \mathbf{s}_t, \mathbf{a}_t).
\label{eq:prob_traj}
\end{align}
To replicate a demonstrator's policy $\pi_{\theta^D}(\mathbf{a}_t^{D} | \mathbf{s}_t)$, the error of a query policy $\pi_{\theta}$ and the demonstrator's policy $\pi_{\theta^D}$ using trajectories $\boldsymbol{\tau}$ is defined as:
\begin{align}
J(\pi_{\theta}, \pi_{\theta^{D}}, \boldsymbol{\tau}) = \sum_{t=1}^{T} \mathbb{E}_{\pi_{\theta}(\mathbf{a}_t | \mathbf{s}_t), \pi_{\theta^{D}}(\mathbf{a}_t^{D} | \mathbf{s}_t)}\left[\left\|\mathbf{a}_t-\mathbf{a}_t^{D}\right\|_{2}^{2}\right].
\label{eq:bc_objective}
\end{align}     
To minimize the expected surrogate loss as the objective function, the parameter ${\theta}^R$ of a BC policy $\pi_{\theta^R}(\mathbf{a}_t^{R} | \mathbf{s}_t)$ is acquired by solving:
\begin{align}
\theta^{R}=\argmin_{\theta} \mathbb{E}_{p(\boldsymbol{\tau} | \pi_{\theta^{D}})}\left[J(\pi_{\theta}, \pi_{\theta^{D}}, \boldsymbol{\tau})\right].
\label{eq:bc_policy}
\end{align}
However, error compounding caused by limited variation in demonstrations can result in policies learned by BC being unstable \cite{ross2010efficient}. The difference between the trajectory distribution of learned policy $\pi_{\theta^R}$ and the demonstrator's policy $\pi_{\theta^D}$ can be formalized as the covariate shift:
\begin{align}
\left|\mathbb{E}_{p(\boldsymbol{\tau} | \pi_{\theta^{D}})}\left[J(\pi_{\theta^{R}}, \pi_{\theta^{D}}, \boldsymbol{\tau})\right]-\mathbb{E}_{p(\boldsymbol{\tau} |\pi_{\theta^{R}})}\left[J(\pi_{\theta^{R}}, \pi_{\theta^{D}}, \boldsymbol{\tau})\right]\right|.
\label{eq:cov_shift}
 \end{align}
In the context of executing the learned policy, a high covariate shift tends to lead policies into undemonstrated states, where it is hard to recover back to a desired trajectory.

\subsection{Disturbances for Augmenting Robot Trajectories (DART) }
\label{s:dart}

To mitigate covariate shift, DART\cite{laskey2017dart} learns policies that are robust to error compounding by generating demonstrations with disturbance injection. The learning space is thereby augmented with recovery actions, allowing for greater variation in demonstration trajectories. 
The level of disturbance is optimized and updated iteratively during data collection over $K$ iterations (number of policy updates), on the disturbance distribution parametrized by $\psi$.

Initially, disturbance injected policy given the disturbance parameter $\psi$ is expressed as $\pi_{\theta}(\mathbf{a}_t | \mathbf{s}_t, \psi)$. By substituting this policy into the trajectory distribution (\equref{eq:prob_traj}), the disturbance injected trajectory distribution is defined as:
\begin{align}
p(\boldsymbol{\tau} | \pi_{\theta}, \psi) = p(\mathbf{s}_1) \prod_{t=1}^{T} \pi_{\theta}(\mathbf{a}_t | \mathbf{s}_t, \psi) p(\mathbf{s}_{t+1} | \mathbf{s}_t, \mathbf{a}_t).
\label{eq:prob_traj_noise}
\end{align}
As the covariate shift (\equref{eq:cov_shift}) cannot be computed explicitly, DART employs the upper bound of the covariate shift derived by applying Pinsker's inequality:
\begin{align}
&\left|\mathbb{E}_{p(\boldsymbol{\tau} | \pi_{\theta^{D}}, \psi)}\left[J(\pi_{\theta^{R}}, \pi_{\theta^{D}}, \boldsymbol{\tau})\right]-\mathbb{E}_{p(\boldsymbol{\tau} | \pi_{\theta^{R}})} \left[J(\pi_{\theta^{R}}, \pi_{\theta^{D}}, \boldsymbol{\tau})\right]\right| \nonumber \\
&\leq T \sqrt{\frac{1}{2} \mathrm{KL}\left(p(\boldsymbol{\tau} | \pi_{\theta^{R}}) || p(\boldsymbol{\tau} | \pi_{\theta^{D}}, \psi)\right)}, 
\label{eq:pinskers}
\end{align}
where, $\mathrm{KL}(\cdot\| \cdot)$ is Kullback-Leibler divergence.
By canceling common factors in trajectory distributions (\equref{eq:prob_traj} and \equref{eq:prob_traj_noise}), the KL divergence can be expanded as follows:
\begin{align}
&\mathrm{KL}\left(p(\boldsymbol{\tau} | \pi_{\theta^{R}}) || p(\boldsymbol{\tau} | \pi_{\theta^{D}}, \psi)\right) \nonumber \\
&= \mathbb{E}_{p(\boldsymbol{\tau} | \pi_{\theta^{R}})} \sum_{t=1}^{T} \log \frac{\pi_{\theta^R}(\mathbf{a}_t^{R}|\mathbf{s}_t)}{\pi_{\theta^D}(\mathbf{a}_t^{R}|\mathbf{s}_t, \psi)}.
\label{eq:kld}
\end{align}
Then, the disturbance parameter $\psi$ is optimized to minimize the upper bound of covariate shift as:
\begin{align}
\min_{\psi} \mathbb{E}_{p(\boldsymbol{\tau} | \pi_{\theta^{R}})} \left[ - \sum_{t=1}^{T} \log \left[ \pi_{\theta^D}(\mathbf{a}_t^{R}|\mathbf{s}_t, \psi) \right] \right].
\label{eq:cov_min}
\end{align}
The expectation of the trajectory distribution with a learned policy $p(\boldsymbol\tau|\pi_{\theta^R})$ cannot be solved since the learned policy's trajectories are not directly observed. Therefore, DART solves \equref{eq:cov_min} using the demonstrator's trajectory with the $k$-th updated disturbance parameter $p\left(\boldsymbol{\tau} | \pi_{\theta^{D}}, \psi_{k}\right)$ instead of the learned policy's trajectory distribution $p\left(\boldsymbol{\tau} | \pi_{\theta^R}\right)$ as:
\begin{align}
&\psi_{k+1} = \argmin_{\psi} \mathbb{E}_{p(\boldsymbol{\tau}|\pi_{\theta^{D}}, \psi_{k})} \left[L(\psi, \pi_{\theta^{R}}^{k}, \pi_{\theta^{D}}, \boldsymbol\tau)\right],
\label{eq:dart_disturbance} \\
&L(\psi, \pi_{\theta^{R}}, \pi_{\theta^{D}}, \boldsymbol\tau) = - \sum_{t=1}^{T} \mathbb{E}_{\pi_{\theta^R}(\mathbf{a}_t^{R}|\mathbf{s}_t)}  \log \left[ \pi_{\theta^D}(\mathbf{a}_t^{R}|\mathbf{s}_t, \psi) \right], 
\label{eq:objective:disturbance}
\end{align}
where $L$ is an objective function for disturbance optimization. From \equref{eq:bc_policy}, the parameter ${\theta}^R$ of a DART policy $\pi_{\theta^R}^k$ is optimized using the trajectory distribution with disturbance $p(\boldsymbol\tau|\pi_{\theta^D},\psi_k)$ as:
\begin{align}
{\theta}^{R} = \argmin_{\theta} \sum_{i=1}^{k} \mathbb{E}_{p\left(\boldsymbol{\tau} | \pi_{\theta^{D}}, \psi_{i}\right)} J\left(\pi_{\theta}, \pi_{\theta^{D}}, \boldsymbol{\tau}\right).
\label{eq:dart_policy}
\end{align}
Note that, a policy $\pi_{\theta^{R}}^{k}$ is learned using all data up to the $k$-th iteration, however, a disturbance parameter $\psi_{k+1}$ is learned using only demonstration data with the most recent $\psi_k$. 

A key limitation of DART is that \equref{eq:pinskers} assumes the demonstrator's policy with disturbance injection performs optimally to induce the distribution of learned trajectories closer to the demonstrations. However, if the demonstration data is diverse-quality, \naive application of DART (\ie optimizing the disturbance parameter $\psi$ over the entire space of trajectories), means that the learned policy cannot be induced towards the high reward demonstrations, and learned DART policies fail to replicate an optimal demonstrator's behavior.

%%%%%%%%%%%%%%%%%%%%%%%%%%%%%%%%%%%%%%%%%%%%%%%%%%%%%%%%%%%%%%%%%%%%%%%%%%%%%%%%
\section{PROPOSED METHOD}
In this section, a novel imitation learning framework is introduced, \textit{task achievement weighted disturbance injections (TAW-DI)}, that learns a robust optimal policy by exploiting task achievement from diverse-quality demonstration data. In contrast to \naive DART, this method generates disturbances as controlled by the task achievement, thereby inducing the learned policy towards the high task achievement demonstrations. The proposed method is outlined in \algorithmref{taw-di}.

\begin{algorithm}[t]
    \footnotesize
	\SetAlgoLined
	\DontPrintSemicolon
    \For{$k = 0$ to $K$}{
        \For{$e = 0$ to $E$}{
            Collect disturbance injected augmented trajectories with task achievement: $\boldsymbol{\tau}_{k}^e \sim  p(\boldsymbol{\tau},o=1 \mid \pi_{\theta^{D}}, \psi_{k})$\;
        }
        policy $\pi_{\theta^{R}}^{k}$ is set with \equref{eq:rw_dart_policy}\;
        disturbance $\psi_{k+1}$ is set with \equref{eq:rw_dart_disturbance}\;
    }
    \caption{\footnotesize TAW-DI}
    \label{algorithm:taw-di}
\end{algorithm}

% \begin{algorithm}[tb]
% \caption{\footnotesize TAW-DI}
% \label{algorithm:taw-di}
% \begin{algorithmic}[1]
% \footnotesize
% \renewcommand{\algorithmicrequire}{\textbf{Input:}}
% \renewcommand{\algorithmicensure}{\textbf{Output:}}
% \FOR {$k = 0$ to $K$}
% \FOR {$e = 0$ to $E$}
% \STATE Collect disturbance injected augmented trajectories with task achievement: $\boldsymbol{\tau}_{k}^e \sim  p(\boldsymbol{\tau},o=1 | \pi_{\theta^{D}}, \psi_{k})$
% \ENDFOR
% \STATE policy $\pi_{\theta^{R}}^{k}$ is set with \equref{eq:rw_dart_policy}
% \STATE disturbance $\psi_{k+1}$ is set with \equref{eq:rw_dart_disturbance}
% \ENDFOR
% \end{algorithmic} 
% \end{algorithm}

\subsection{TAW-DI}
Initially, a binary optimality variable $o$ indicating the maximum possible task achievement is introduced. Maximizing the task achievement is equivalent to maximizing the likelihood of $o=1$ (optimal) \cite{luck2016sparse}, and this enables learning optimal policies from diverse-quality demonstrations as a probabilistic operation. The conditional probability of receiving maximum task achievement given a cost function is defined as:
\begin{align}
p(o=1|\boldsymbol\tau) = \alpha~\mathrm{exp}(-c(\boldsymbol\tau)),
\label{eq:distribution_optimality_variable}
\end{align}
where, $c(\boldsymbol\tau)$ is any cost function that calculates task achievement and $\alpha$ is a normalizing constant.
By using this probability into \equref{eq:prob_traj_noise}, the task achievement weighted trajectory distribution collected with disturbance injection is defined as:
\begin{align}
&p(\boldsymbol{\tau},o=1 | \pi_{\theta}, \psi) \nonumber \\ 
&= p(o=1|\boldsymbol\tau) p(\mathbf{s}_1) \prod_{t=1}^{T} \pi_{\theta}(\mathbf{a}_t | \mathbf{s}_t, \psi) p(\mathbf{s}_{t+1} | \mathbf{s}_t, \mathbf{a}_t).
\label{eq:prob_traj_noise_rw}
\end{align}
By substituting this to the trajectory distribution of \equref{eq:dart_disturbance}, the update scheme of disturbance parameter considering the task achievement is given as:
\begin{align}
\psi_{k+1} &= \argmin_{\psi} \mathbb{E}_{p(\boldsymbol{\tau},o=1 | \pi_{\theta^{D}}, \psi_k)} \left[ L(\psi, \pi_{\theta^{R}}^{k}, \pi_{\theta^{D}}, \boldsymbol\tau) \right]
\label{eq:rw_dart_disturbance}.
\end{align}
From \equref{eq:dart_policy}, the update scheme of policy parameter considering the task achievement is given as:
\begin{align}
\theta^{R} = \argmin_{\theta} \sum_{i=1}^{k} \mathbb{E}_{p(\boldsymbol{\tau},o=1 | \pi_{\theta^{D}}, \psi_{i})}\left[J\left(\pi_{\theta}, \pi_{\theta^{D}}, \boldsymbol{\tau}\right)\right].
\label{eq:rw_dart_policy}
\end{align}

\subsection{TAW-DI with deep neural network policy model}
As an implementation of TAW-DI, a deterministic deep neural network model is employed as the policy $\pi_{\theta^{NN}}(\mathbf{s}_t)$ and the disturbance distribution follows a Gaussian noise $\mathcal N(0, \Sigma)$ parametrized by $\Sigma$. By substituting the policy $\pi_{\theta^{NN}}(\mathbf{s}_t)$ and disturbance parameter $\Sigma$ into the objective function $L$ (\equref{eq:objective:disturbance}) of disturbance (\equref{eq:rw_dart_disturbance}), we obtain the disturbance optimization as:
\begin{align}
\Sigma_{k+1} &= \argmin_{\Sigma} \mathbb{E}_{p(\boldsymbol{\tau},o=1 | \pi_{\theta^{D}}, \Sigma_k)} \nonumber \\
&\left[ - \sum_{t=1}^{T} \log \mathcal{N} \left(\pi_{\theta^{NN}}(\mathbf{s}_t) | \mathbf{a}_t^{D}, \Sigma \right) \right]
\end{align}
By applying the Monte Carlo method to the expectation over trajectory distribution, the solution is approximated as: 
\begin{align}
\hat{\Sigma}_{k+1} & = \frac{1}{ET} \sum_{e=1}^{E} \sum_{t=1}^{T} c(\boldsymbol{\tau}_{k}^e) \nonumber \\
& \left[ \left(\pi_{\theta^{NN}}(\mathbf{s}_t) - \mathbf{a}_t^{D} \right)\left(\pi_{\theta^{NN}}(\mathbf{s}_t) - \mathbf{a}_t^{D} \right)^{T} \right]
\label{eq:loss_dart}
\end{align}
where $E$ is a total episodes (number of trajectories).
The policy parameter optimization of  \equref{eq:rw_dart_policy} is solved by applying the gradient descent methods.

\begin{figure}[tb]
\centering
\includegraphics[width=0.9\hsize] {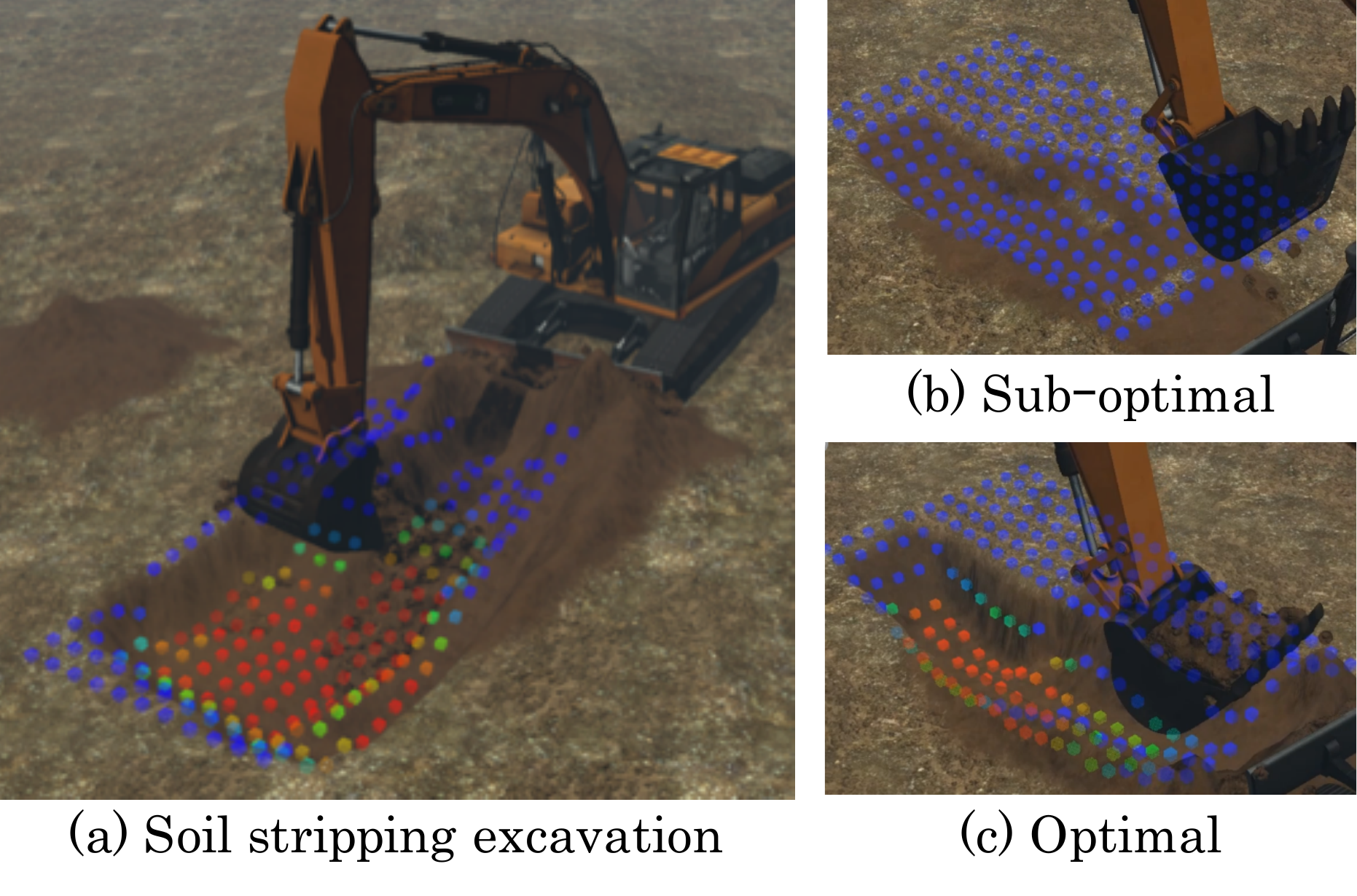}
\caption{Simulation environment on VortexStudio. Intensity of color indicates depth of excavations.}
\label{fig:experiment_simulation_task1}
\end{figure}

%%%%%%%%%%%%%%%%%%%%%%%%%%%%%%%%%%%%%%%%%%%%%%%%%%%%%%%%%%%%%%%%%%%%%%%%%%%%%%%%
\section{EVALUATION}
To investigate the effectiveness of learning robust policies given \textit{limited variation} and \textit{diverse-quality} demonstration data, an autonomous excavation task is presented in both simulation and a scale robotic excavator. Specifically, a soil stripping excavation task is performed, where an area is cleared of soil by three consecutive scooping motions, as shown in \figref{experiment_simulation_task1}(a). 

Excavation is a dangerous and strenuous activity for operators, and robotic automation is actively researched \cite{dadhich2016key}, with previous studies \cite{bradley1998development, ha2002robotic} investigating automation via modeling of excavator kinematics, or learning data-driven policies using deep-learning \cite{fukui2017imitation, son2020expert, egli2020towards}. RL is unsuitable to this task, due to the danger of exploration, and previous studies have investigated imitating the human behavior for automation \cite{fukui2017imitation, son2020expert}. This paper addresses the challenging task of learning from diverse-quality demonstrations, which is a natural consequence of the difficulty of excavation.

\begin{figure}[tb]
\centering
\includegraphics[width=1.0\hsize] {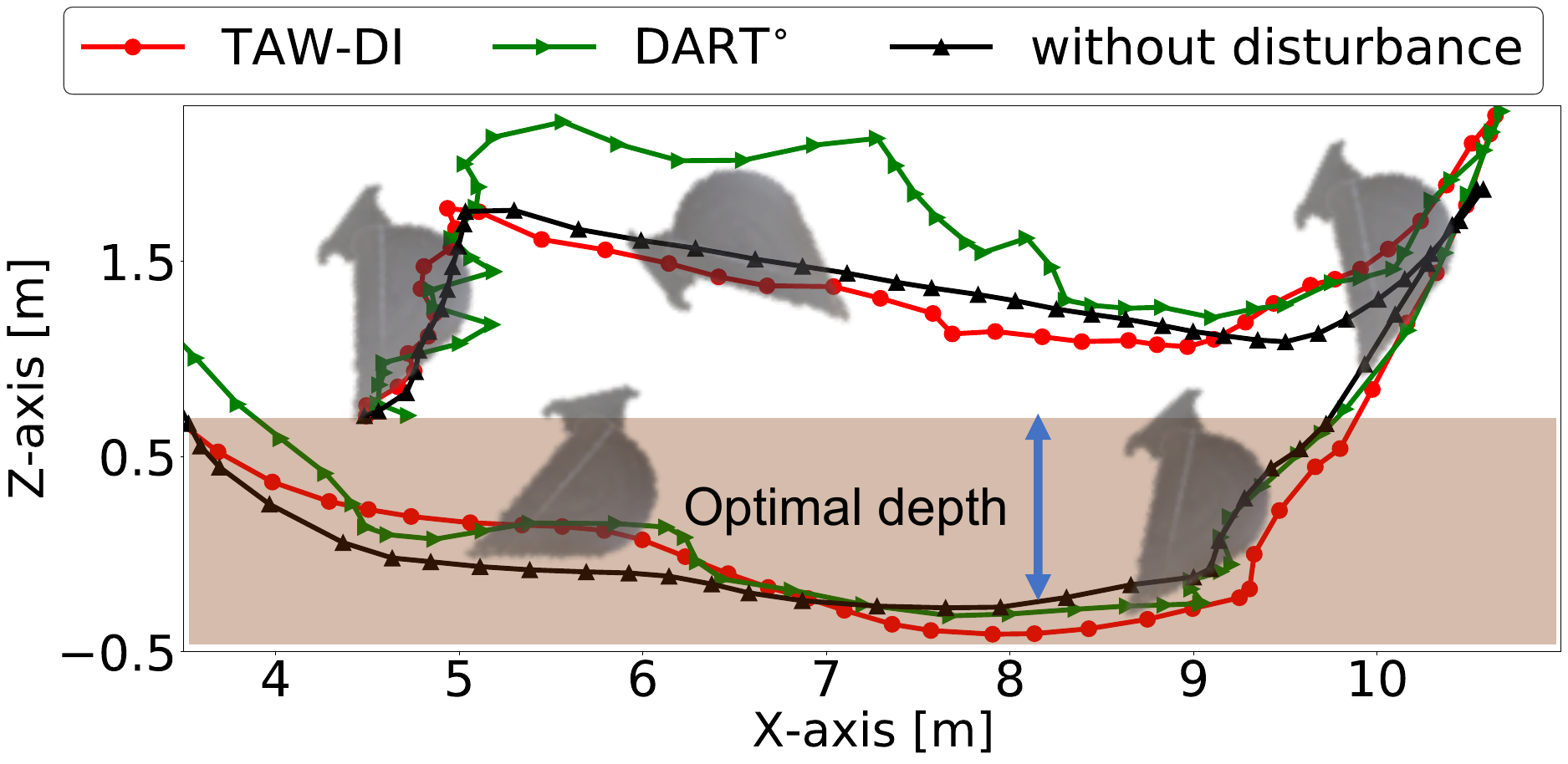}
\caption{
Visualization of environment, showing excavator's bucket position (gray shaded) and soil area (brown shaded, with topsoil starting at 0.6m). Results of demonstration trajectories are plotted, either without disturbances (BC), or with disturbances (TAW-DI, and DART$^{\circ}$).}
\label{fig:experiment_simulation_task1_demonstration_traj}
\end{figure}

\begin{figure}[tb]
\centering
\includegraphics[width=0.9\hsize] {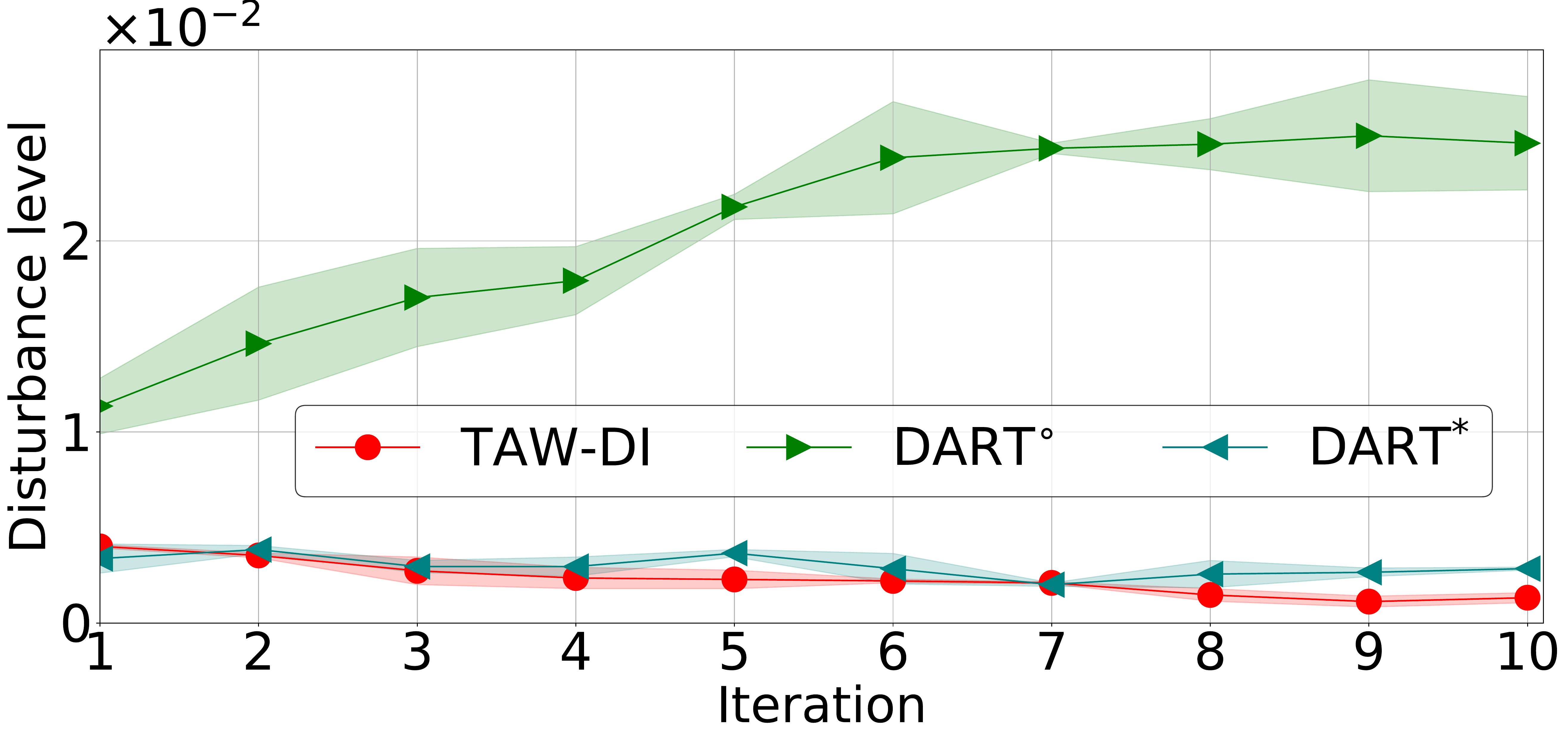}
\caption{Result of disturbance update by \equref{eq:dart_disturbance} and \equref{eq:rw_dart_disturbance}. TAW-DI and DART$^{\circ}$ learn from sub-optimal trajectories, while DART$^*$ removes them and updates its disturbance using the optimal trajectory.}
\label{fig:experiment_simulation_task1_noise}
\end{figure}

\begin{figure*}[tb]
    \centering
    \begin{minipage}[b]{0.22\linewidth}
    \includegraphics[width=1.0\hsize]{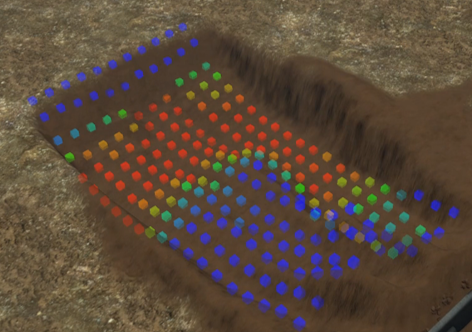}
    \subcaption{TAW-DI}
    \end{minipage}
    \begin{minipage}[b]{0.22\linewidth}
    \includegraphics[width=1.0\hsize]{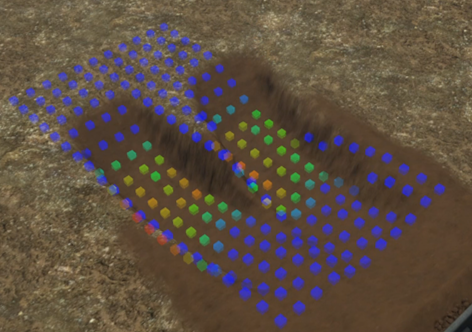}
    \subcaption{TAW-BC}
    \end{minipage}
    \begin{minipage}[b]{0.22\linewidth}
    \includegraphics[width=1.0\hsize]{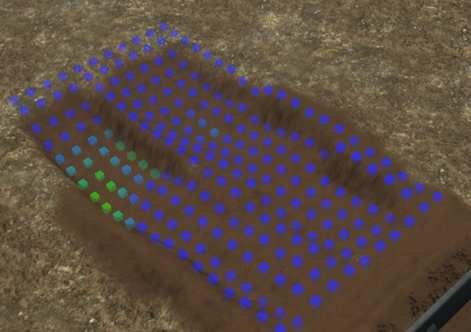}
    \subcaption{DART$^{\circ}$}
    \end{minipage}
    \begin{minipage}[b]{0.22\linewidth}
    \includegraphics[width=1.0\hsize]{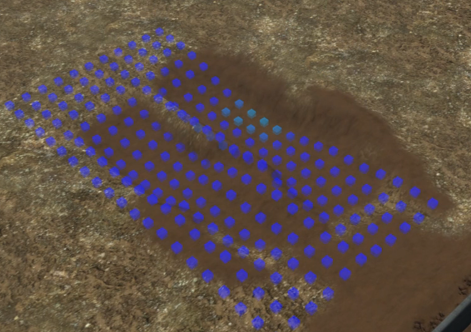}
    \subcaption{BC$^{\circ}$}
    \end{minipage}
    \caption{Soil stripping excavation task by learned policies. TAW-DI obtained the best result close to the optimal trajectory by applying disturbance injection and task achievement weighting. TAW-BC failed to excavate the proper position due to the error compounding. DART$^{\circ}$ performed soil stripping robustly, but it is shallow due to the effect of learning sub-optimal trajectories. BC$^{\circ}$ was affected by both problems, and the result is worse.}
    \label{fig:experiment_simulation_task1_test_result}
    \vspace{-4mm}
\end{figure*}

\begin{figure}[tb]
\centering
\includegraphics[width=0.9\hsize] {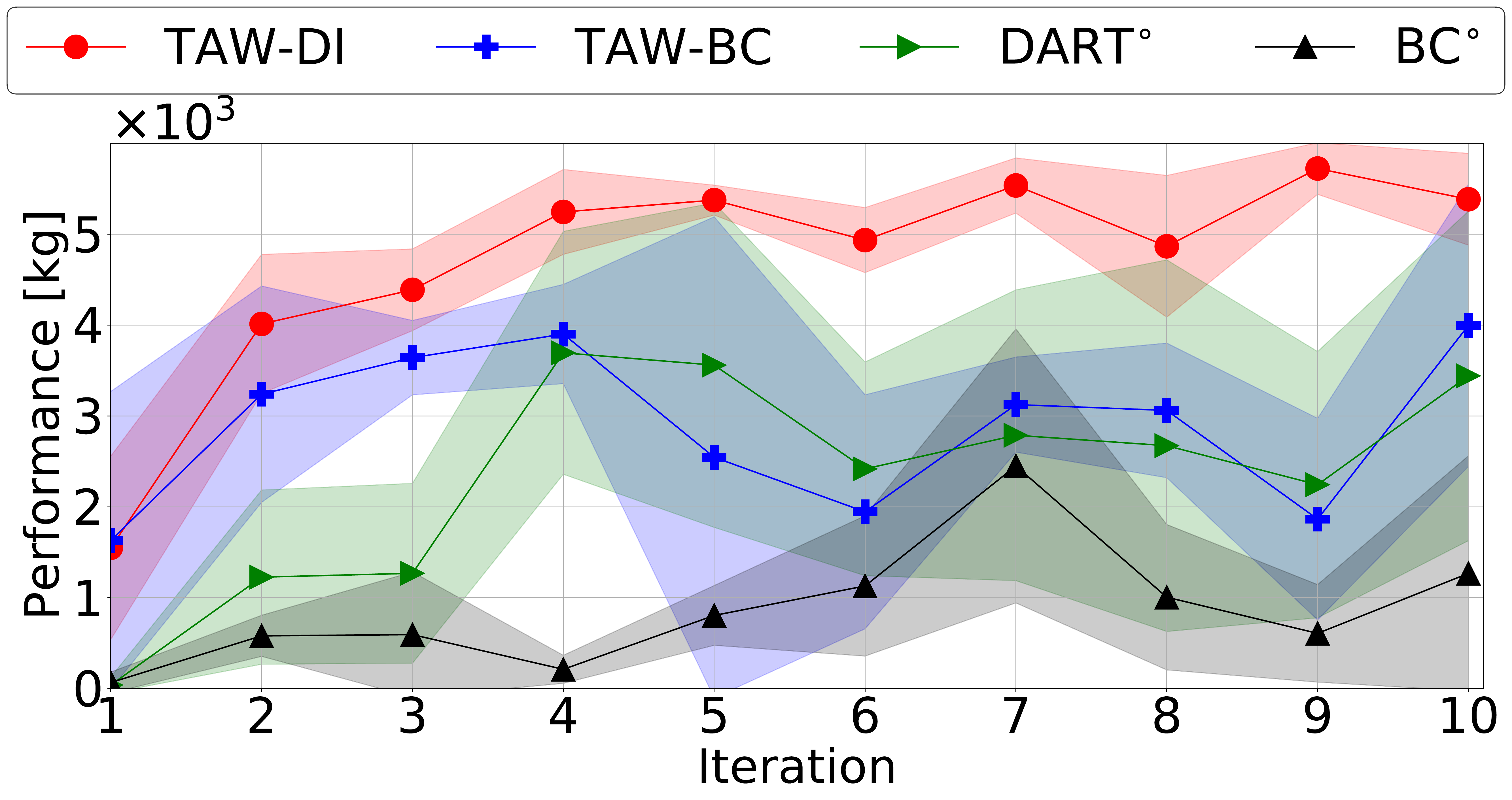}
\caption{Test performance. All method utilizes both optimal and sub-optimal trajectories. While the performance of the comparisons, TAW-BC, DART$^{\circ}$, and BC$^{\circ}$, were deteriorated due to the limited variation and diverse-quality of demonstrations, TAW-DI showed the best performance.}
\label{fig:experiment_simulation_task1_performance}
\end{figure}

%%%%%%%%%%%%%%%%%%%%%%%%%%%%%%%%%%%%%%%%%%%%%%%%%%%%%%%%%%%%%%%%%%%%%%%%%%%%%%%%
\subsection{Simulation experiment}
\label{s:sim}
To evaluate the proposed TAW-DI method, comparisons are made between baselines BC and DART, and TAW-BC (task achievement weighting without disturbance injections) is used as an ablation study. Generally, BC and DART assume demonstrations contain only optimal trajectories defined via a strict threshold, and these are denoted as BC$^*$ and DART$^*$. However, in general, choosing this threshold \textit{a priori} can be challenging, and as such BC$^{\circ}$ and DART$^{\circ}$ are additional comparisons that also utilize both optimal and sub-optimal trajectories. The simulation environment is developed on the VortexStudio simulator \cite{noauthor_2019-dx} which enables the real-time simulation of excavation with soil dynamics.

\subsubsection{Task Setting}
The diverse-quality of the demonstration data is defined according to domain knowledge \cite{SafeWork_undated-zf}:
\begin{enumerate*}
\item Deep excavation ($>1.5~m$) of the ground should be avoided as it may cause trench or excavator instability.
\item Shallow excavation ($<0.5~m$) (\figref{experiment_simulation_task1}(b)) is considered sub-optimal, as it necessitates additional work and time to clear the soil.
\item The appropriate depth ($\sim1.0~m$) of excavation is considered as the optimal trajectories (\figref{experiment_simulation_task1}(c)).
\end{enumerate*} 
A custom controller is designed that automatically performs soil striping, either optimally or sub-optimally. The controller automates the excavator to emulate the excavation trajectory that follows the idea of previous work \cite{jud2019autonomous}, where the excavation motion is divided into several simple trajectories, by passing through a set of predefined points (known as \textit{support points}). To make the demonstration more diverse, uniform noise $\mathcal{U}(-\sigma, \sigma)$ is added to the support points. In this experiment, $\sigma = 0.3~\mathrm{m}$ is chosen as a reasonable divergence, as this task involves a large spatial scale of the excavator and work-space. The uniform noise is also added to the initial robot position as task uncertainty.

The state is described by 33 dimensions, these being
\begin{enumerate*}
\item the excavator's state, consisting of the joint angle and angular velocity of four joints,
\item the soil's shape information, as seen by an overhead depth camera, consisting of a vectorized $5 \times 5$ grid of the excavation space,  used to determine the next excavation point by considering the soil shape.
\end{enumerate*}
The action is the velocity of each joint. Since it is difficult to measure the exact depth of the excavated soil in a real situation, this task defines the task achievement ($c(\boldsymbol\tau)$ in \equref{eq:distribution_optimality_variable}) as the total amount of moved soil which can easily be evaluated at the end of the task, and it is suitable to our method as discussed in \sref{rl}.

Following the DART architecture \cite{laskey2017dart} with the same parameterization, a four-layer deep neural network consisting of an input layer, two 64-units hidden layers, and an output layer is used to train the policies. In the demonstration phase, demonstration data is collected for $10$ iterations of $2$ episodes (as in DART \cite{laskey2017dart}), and data is used to update the policies and disturbances. Each iteration contains an optimal and a sub-optimal trajectory. Variations in demonstrations may induce performance disparity in learned policies so that the demonstration phase is repeated twice to validate the robustness. In the testing phase, each of the policies learned in each iteration is evaluated three times. The one episode takes about 600 steps to complete the task.

\begin{figure}[tb]
\centering
\includegraphics[width=0.9\hsize] {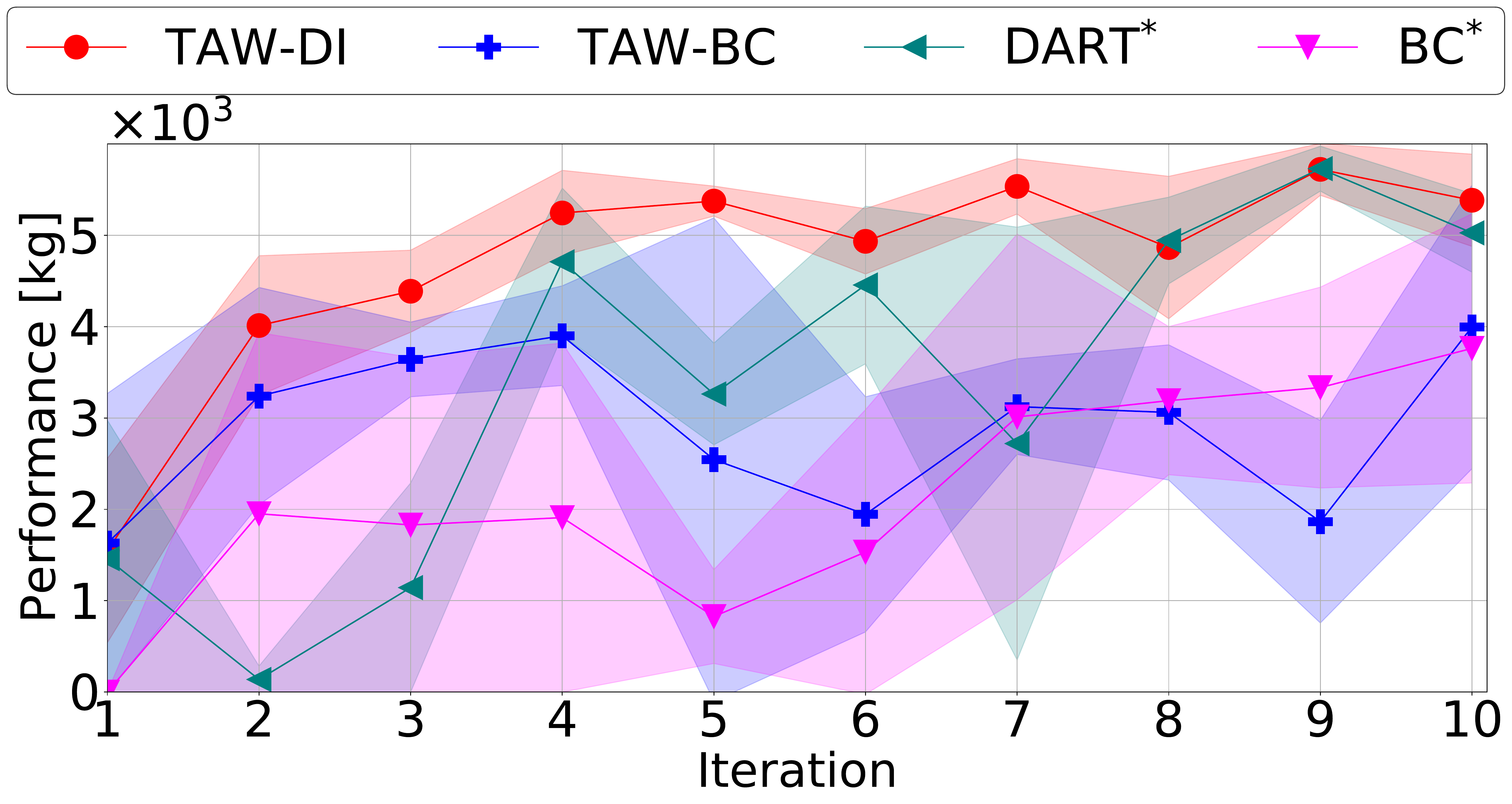}
\caption{Test performance. TAW-DI and TAW-BC utilize an optimal and a sub-optimal trajectory, and DART$^*$ and BC$^*$ only use an optimal trajectory. The performance convergence of TAW-DI and TAW-BC is faster than DART$^*$ and BC$^*$.}
\label{fig:experiment_simulation_task1_performance_optimal}
\end{figure}

\subsubsection{Results}
The demonstration trajectories collected from the last iteration are shown in \figref{experiment_simulation_task1_demonstration_traj}. Initially, DART$^{\circ}$ exhibits dangerous oscillation and undesired drift due to the diverse-quality demonstrations in \figref{experiment_simulation_task1_demonstration_traj}, causing the disturbance level to continuously grow over training iterations, as seen in \figref{experiment_simulation_task1_noise}. In contrast, DART$^*$, which only uses optimal trajectories, does not experience this phenomenon. In comparison, TAW-DI generates demonstration trajectories that are more stable and do not drift in \figref{experiment_simulation_task1_demonstration_traj}. TAW-DI prevents excessive disturbance level updates by the weighting of trajectories, as is shown in \figref{experiment_simulation_task1_noise} where the disturbance level converges.

After policy learning, learned policies are evaluated, as seen in \figref{experiment_simulation_task1_test_result}, with performance seen in  \figref{experiment_simulation_task1_performance}. Initially, the proposed method, TAW-DI, which applies both disturbance injection and task achievement weighting, performs the best and achieves a performance close to the optimal. In contrast, it is seen that BC$^{\circ}$ performs poorly, due to the fact it suffers from both error compounding and diverse-quality of demonstration data, resulting in undesired behavior such as excavating improper position. DART$^{\circ}$ performs better than BC, as it explicitly accounts for error compounding by robustification using the disturbance injection. However, it is similarly unable to deal with the problem of diverse-quality demonstrations, resulting in sub-optimal policy learning. TAW-BC is robust against the problem of a diverse-quality demonstration by task achievement weighting, but it is not stable due to the error compounding compared with TAW-DI.

In addition, there are clear benefits to utilizing all demonstrations (optimal and sub-optimal), as seen in \figref{experiment_simulation_task1_performance_optimal}. BC$^*$ fails to reach optimal performance even when only using optimal demonstrations, due to the aforementioned inability to deal with error compounding (similarly seen with TAW-BC). DART$^*$ overcomes this, and reaches optimal performance, however, at the cost of a slow convergence rate due to the limited number of optimal samples. In comparison, TAW-DI, which additionally utilizes sub-optimal trajectories, reaches this optimal level of performance much faster. As such, the proposed task achievement weighting method is not only more generalizable, as there is no need to define an optimal threshold, but is more suitable to data-limited environments.

\begin{figure}[t]
\centering
\includegraphics[width=1.0\hsize] {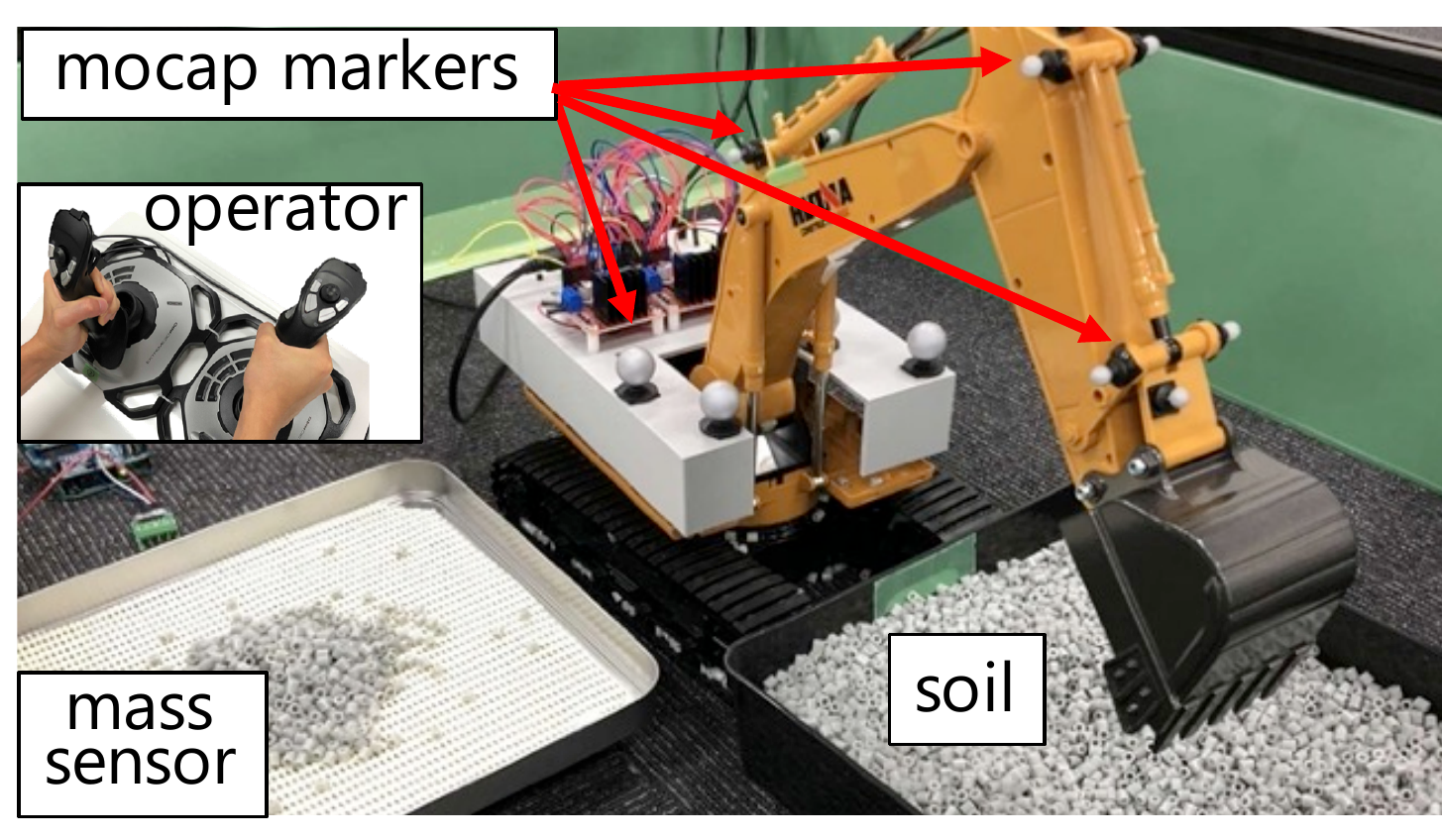}
\caption{A scale robotic excavator environment.}
\label{fig:experiment_realrobot_task1_setting}
\end{figure}

%%%%%%%%%%%%%%%%%%%%%%%%%%%%%%%%%%%%%%%%%%%%%%%%%%%%%%%%%%%%%%%%%%%%%%%%%%%%%%%%
\subsection{Real robot experiment}
In this section, all methods utilize both optimal and sub-optimal demonstrations, including the baselines, BC$^{\circ}$ and DART$^{\circ}$, and the proposed method TAW-DI. A scale robotic excavator is used for this experiment as shown \figref{experiment_realrobot_task1_setting}.

\subsubsection{Task setting}
This experiment employs human demonstrators. To verify the effectiveness of the proposed method, the demonstrators are instructed to collect an optimal and a sub-optimal trajectories in each episode to keep the balance of the quality of demonstrations. The additional variation in diverse-quality demonstrations is induced by the uncertainty of the demonstrator's operation instead of the uniform noise used in the simulation. The demonstrators manipulate the excavator using twin joystick controllers where joints are controlled independently.

State, action, and task achievement are consistent with \sref{sim}, and a motion capture system (OptiTrack Flex3) detects markers on each arm of the excavator and estimates the joint angles and angular velocities simultaneously. A depth camera (Intel RealSense D455) attached overhead captures soil depth. A mass sensor measures the total soil mass. Plastic beads are used instead of real soil, corresponding to soil dynamics with low shear strength and high flow velocity. 

In the demonstration phase, demonstration data is collected for $5$ iterations of $2$ episodes (an optimal and a sub-optimal) by demonstrators, specifically, by two subjects. The policies are learned by the same network with \sref{sim}. In the testing phase, the learned policies from the final iteration are evaluated three times. One episode takes 200 steps to complete the task.

\subsubsection{Results}
\begin{figure}[tb]
    \centering
    \begin{minipage}[b]{0.48\linewidth}
    \includegraphics[width=1.0\hsize]{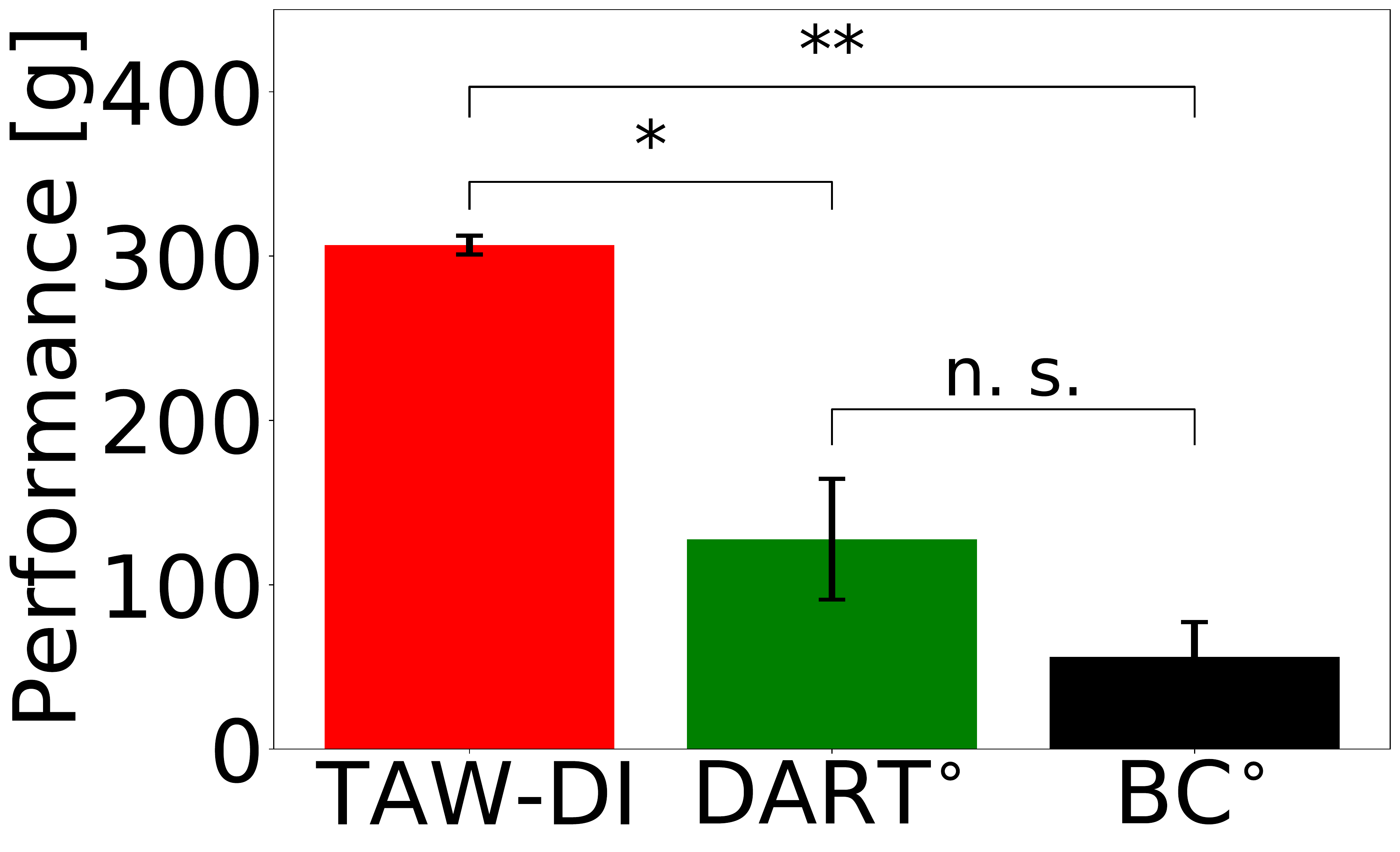}
    \subcaption{Subject 1}
    \end{minipage}
    \begin{minipage}[b]{0.48\linewidth}
    \includegraphics[width=1.0\hsize]{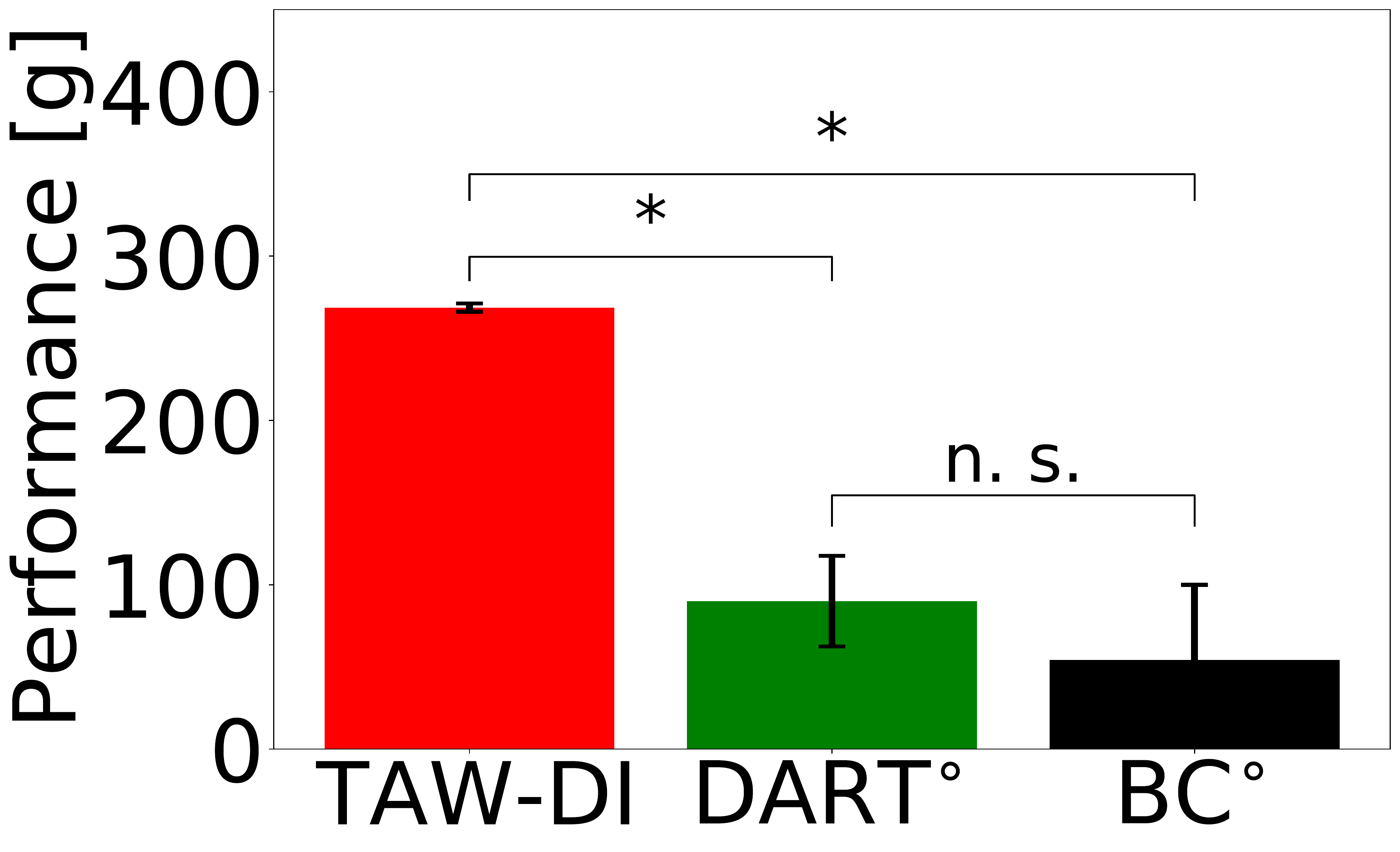}
    \subcaption{Subject 2}
    \end{minipage}
    \caption{Test performance of two subjects. Significant differences by t-test were observed between the proposed method and baselines ($*:p < 0.05, **:p < 0.005$).}
    \label{fig:experiment_realrobot_task1_test}
\end{figure}

The performance of the test trajectories by two subjects is shown in \figref{experiment_realrobot_task1_test}, where it is seen that as expected, BC$^{\circ}$ performs poorly due to error compounding and the diverse-quality demonstrations, and the diverse-quality demonstrations also deteriorate the performance of DART$^{\circ}$. In contrast, TAW-DI obtains the same performance as the optimal trajectories by the disturbance injection and the task achievement weighting similar to the simulation experiments. Such results are similarly obtained for two subjects, and the significant differences by t-test were observed between the proposed method and baselines.

%%%%%%%%%%%%%%%%%%%%%%%%%%%%%%%%%%%%%%%%%%%%%%%%%%%%%%%%%%%%%%%%%%%%%%%%%%%%%%%%
\section{Discussion}
In this section, important discussions for the experimental results are presented.

(1) \textit{What effect does combining disturbance injections with task achievement weighting have on policy learning?} When demonstrations contain sub-optimal trajectories, DART is unstable due to the disturbance update continuously growing (\figref{experiment_simulation_task1_noise}) in an attempt to apply disturbances that minimize the difference between the optimal and sub-optimal trajectories. However, this large disturbance adversely deteriorates the performance of the demonstration and, consequently, the policy's performance during testing. Under the assumption of diverse-quality demonstrations as common in real-world problems, combining disturbance injection with task achievement weighting overcomes the issues by focusing on mainly learning from high task achievement demonstrations.

(2) \textit{What is the advantage of utilizing weighted sub-optimal trajectories instead of simply eliminating them?} Experimental results show the use of sub-optimal trajectories can accelerate the convergence of the policy performance. In real-world scenarios where the data collection cost is high due to the complexity of the operation and use of long-term tasks, the concept of weighting and utilizing a small number of demonstration trajectories without removing them is significant. This concept is applicable to various real-world robotics problems like excavation, which demands a high cost in human operation and the use of real robots.

%%%%%%%%%%%%%%%%%%%%%%%%%%%%%%%%%%%%%%%%%%%%%%%%%%%%%%%%%%%%%%%%%%%%%%%%%%%%%%%%
\section{Conclusion}
This paper presents a novel imitation learning framework for addressing real-world robotics problems that suffer from the dual problem of \textit{limited variation} and \textit{diverse-quality} of demonstrations. While previous studies have investigated these problems independently, the proposed method consistently outperforms methods that explicitly address these problems independently in both simulation and real robot experiments. As future works, the scalability of this study can be improved by making the policy multi-modal \cite{sasaki2021variational, Oh2021} for further applicability to more complex tasks.

%%%%%%%%%%%%%%%%%%%%%%%%%%%%%%%%%%%%%%%%%%%%%%%%%%%%%%%%%%%%%%%%%%%%%%%%%%%%%%%%
\bibliographystyle{IEEEtran}
\bibliography{reference}

\end{document}